\documentclass{article} 
\usepackage{iclr2026_conference,times}
\usepackage{graphicx} 

\usepackage{amsmath,amsfonts,bm}

\def\eqref#1{equation~\ref{#1}}

\def\1{\bm{1}}

\DeclareMathAlphabet{\mathsfit}{\encodingdefault}{\sfdefault}{m}{sl}
\SetMathAlphabet{\mathsfit}{bold}{\encodingdefault}{\sfdefault}{bx}{n}

\usepackage[dvipsnames]{xcolor}

\usepackage{hyperref}
\usepackage{url}
\usepackage{placeins}
\usepackage{float}
\usepackage{multirow}
\usepackage{booktabs}
\usepackage{tabularx}
\usepackage[most]{tcolorbox}
\tcbset{colback=gray!3,colframe=gray!70!black,boxrule=0.4pt,arc=1.2pt,left=4pt,right=4pt,top=4pt,bottom=4pt}
\newtcblisting{promptbox}[1]{title=#1,listing only,breakable,fonttitle=\bfseries\footnotesize,listing options={basicstyle=\ttfamily\footnotesize,breaklines=true,columns=fullflexible}}

\title{CraniMem: Cranial Inspired Gated and Bounded Memory for Agentic Systems}
\iclrfinalcopy

\author{Pearl Mody\textsuperscript{1}, Mihir Panchal\textsuperscript{1}, Rishit Kar\textsuperscript{1}, Kiran Bhowmick\textsuperscript{1}, Ruhina Karani\textsuperscript{1}  \\
\textsuperscript{1}Dwarkadas Jivanlal Sanghvi College of Engineering\\
\small
\url{modypearl05@gmail.com}, \url{mihirpanchal5400@gmail.com}, \url{emailrishitkar@gmail.com}\\
\url{kiran.bhowmick@djsce.ac.in}, \url{ruhina.karani@djsce.ac.in}
}

\begin{document}

\maketitle

\begin{abstract}
Large language model (LLM) agents are increasingly deployed in long running workflows, where they must preserve user and task state across many turns. Many existing agent memory systems behave like external databases with ad hoc read/write rules, which can yield unstable retention, limited consolidation, and vulnerability to distractor content. We present \textbf{CraniMem}, a neurocognitively motivated, gated and bounded multi-stage memory design for agentic systems. CraniMem couples goal conditioned gating and utility tagging with a bounded episodic buffer for near term continuity and a structured long-term knowledge graph for durable semantic recall. A scheduled consolidation loop replays high utility traces into the graph while pruning low utility items, keeping memory growth in check and reducing interference. On long horizon benchmarks evaluated under both clean inputs and injected noise, CraniMem is more robust than a Vanilla RAG and Mem0 baseline and exhibits smaller performance drops under distraction. Our code is available at \url{https://github.com/PearlMody05/Cranimem} and the accompanying PyPI package at \url{https://pypi.org/project/cranimem}.
\end{abstract}

\section{Introduction}
Large language model LLM agents increasingly operate over long horizons, they plan, execute tools, collaborate with other agents, and interact with users over days or weeks. In these settings, memory is no longer a simple context window, but a system component that must decide what to store, when to retrieve it, and how to keep knowledge consistent across multiple agent roles and tasks. Recent work has proposed long-term memory stacks for agents, for example production oriented persistent stores and indexing pipelines, as well as specialized agentic memory modules for retrieval and reasoning \citep{chhikara2025mem0,xu2025mem,huang2025licomemory,hu2025memory}. However, many current designs treat memory as an external database with heuristic write/read policies, which can lead to brittle retention where important events are overwritten by noise, weak consolidation across time, and poor synchronization when multiple agents contribute partially overlapping beliefs \citep{yuen2025intrinsic,jiang2025personamem}.

Motivated by findings from neuroscience and cognitive science, we argue that agentic memory should be designed as a 
gated, bounded, and multi-stage process, similar to how biological systems regulate encoding, consolidation, and retrieval across hippocampal cortical pathways \citep{yang2025review,dong2025towards}. Building on this perspective, we introduce \textbf{CraniMem}, a neurobiologically inspired memory framework for agentic systems that explicitly separates transient working and episodic traces, consolidated long-term knowledge, and control gates that decide what enters memory, what is rehearsed, and what is exposed to downstream reasoning.

Our key contributions are:
\begin{enumerate}
    \item We propose \textbf{CraniMem}, a gated and bounded multi-stage memory architecture that supports selective encoding, consolidation, and retrieval for long horizon agent behavior.
    \item We introduce a dual-store memory architecture with explicit consolidation pathways that separate episodic and semantic memory, enabling rapid adaptation alongside stable knowledge formation without unbounded growth.
    \item Through comprehensive evaluation on long horizon benchmarks, we demonstrate that selective forgetting via importance weighting and temporal decay outperforms both unlimited retention and aggressive compression, yielding higher information stability and comparable resistance to distraction.
\end{enumerate}

\section{Related Works}

\subsection{Multi Agent Memory Architectures}
A growing body of work studies how to equip LLM agents with persistent memory beyond the prompt window. Systems such as Mem0 focus on production ready long-term memory with scalable storage, retrieval, and update mechanisms \citep{chhikara2025mem0}. A-Mem formalizes agentic memory as a component that supports writing experiences and retrieving them for downstream planning and decision making \citep{xu2025mem}. Lightweight designs like LiCoMemory aim to reduce memory overhead while maintaining long-term reasoning ability \citep{huang2025licomemory}. Broader surveys highlight recurring design axes what to store between facts and experiences, how to index and retrieve, and how to prevent drift or stale information and emphasize the gap between memory as a database and memory as an adaptive cognitive process \citep{hu2025memory,liu2025unifying}.

In multi-agent settings, memory must also address synchronization and heterogeneity. Intrinsic Memory Agents introduce structured contextual memory to coordinate heterogeneous agents and improve collaboration \citep{yuen2025intrinsic}. PersonaMem-v2 targets personalization by learning implicit user personas and maintaining long-term user specific memory, which introduces additional challenges around stability and updating user models over time \citep{jiang2025personamem}. \citep{wheeler2025procedural} argue that procedural memory alone is insufficient for robust agent behavior, motivating richer memory types and control mechanisms .

\subsection{Neuroscience Inspired Memory Components}
Neuroscience inspired approaches increasingly inform how memory should be organized for LLMs and agents. HippoRAG proposes a long-term memory mechanism driven by neurobiological motivation that draws inspiration from hippocampal indexing for retrieval \citep{jimenez2024hipporag}. Other work explores hippocampal inspired multimodal memory to support long understanding of audiovisual events \citep{lin2025hippomm}. Position papers argue that episodic memory is a key missing ingredient for long-term agent coherence and that agents should store causally linked time stamped experiences rather than only distilled facts \citep{pink2025positionepisodicmemorymissing}.

Complementary lines of research examine cognitive memory mechanisms in LLMs and agentic frameworks, including how memory representations might be structured, consolidated, and used for reasoning \citep{shan2025cognitivememorylargelanguage}. Brain inspired multi-memory agent frameworks for embodied learning further motivate separating memory stores and using control policies to regulate encoding and retrieval \citep{lei2025robomemorybraininspiredmultimemoryagentic}. Our work follows this trend, but focuses specifically on gated and bounded memory operations to improve retention and consistency in long horizon agentic systems.

\section{CraniMem}
\label{sec:cranimem}

\begin{figure}[htb]
    \centering
    \includegraphics[width=1\linewidth]{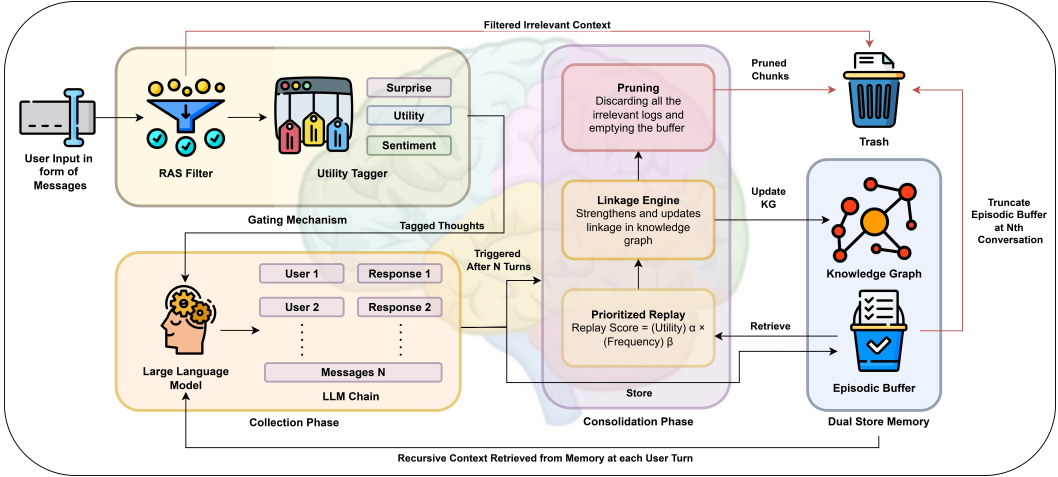}
    \caption{CraniMem architecture. The pipeline implements ingestion via RAS inspired gating and utility tagging, short-term storage in a bounded episodic buffer, optimization through replay selection and pruning to Trash, long-term structural storage via a linkage engine that updates a knowledge graph, and dual-path retrieval using both the buffer and the graph.}
    \label{fig:CraniMem}
\end{figure}

CraniMem is a memory component for long horizon LLM agents designed around two neurocognitive principles. First, attentional gating reduces interference by suppressing low-salience inputs before they reach memory, analogous to brainstem and thalamic control systems that regulate access to downstream processing. Second, systems consolidation stabilizes knowledge by transferring selected experiences from a fast episodic store to a slower semantic store through replay.

Operationally, CraniMem maintains two bounded stores and a periodic optimization loop. Recent interaction traces are stored in an episodic buffer that supports high fidelity short range continuity and provides candidates for consolidation. Durable information such as user preferences, constraints, and stable plans is stored in a structured knowledge graph that supports consistent multi-hop retrieval. A gating module filters candidate writes and assigns utility signals that drive which items are rehearsed. During consolidation, high-utility episodic traces are integrated into the knowledge graph and low utility traces are discarded to keep memory bounded.

Algorithmic details for gating, replay scoring, linkage, and retrieval are provided in Appendix~\ref{appendix:cranimem_details}.

\section{Evaluation Protocol}
\label{sec:protocol}


\paragraph{Clean vs. Noisy evaluation}
In the clean setting, the agent receives only task relevant interaction traces, and the memory write stream contains only those traces. In the noisy setting, we inject distractor snippets into the same write stream to simulate context pollution. Following a fixed injection schedule, we add $m$ irrelevant distractor memories per injection event. The distractors are unrelated to the current goal, but are written to memory using the same write policy as task traces, so they compete with relevant items at retrieval time. We report results under both settings and quantify robustness using the clean and noisy performance gap using the evaluation metrics mentioned in  Appendix~\ref{evaluation_metrics}.

\section{Results}

We evaluate our proposed memory architecture, CraniMem, against Vanilla RAG and Mem0 on Qwen2.5-7B-Instruct. Additionally, to assess generalization across LLMs, we compare CraniMem with Vanilla RAG on multiple instruction-tuned models. For each setup, the underlying LLM and decoding configuration are kept fixed, isolating the impact of the memory system. All models are evaluated under clean inputs and under injected distractor noise of HotpotQA Dataset \citep{yang2018hotpotqa}, following the protocol described in Appendix~\ref{datasets} and using the prompt templates in Appendix~\ref{appendix:prompts}.

\begin{table}[H]
\centering
\small
\renewcommand{\arraystretch}{1.12}
\setlength{\tabcolsep}{4pt}
\begin{tabular}{lcccccccccc}
\toprule

\multirow{2}{*}{\textbf{Architecture}} 
& \multicolumn{4}{c}{\textbf{Noisy}} 
& \multicolumn{4}{c}{\textbf{Clean}} 
& \multirow{2}{*}{\centering\shortstack{\textbf{Noise Drop} \\ \textbf{$\Delta_{\mathrm{noise}}$}}} \\

\cmidrule(lr){2-5} \cmidrule(lr){6-9}

& \textbf{P} & \textbf{R} & \textbf{F1} & \textbf{Latency (s)} 
& \textbf{P} & \textbf{R} & \textbf{F1} & \textbf{Latency (s)} 
& \\

\midrule

Vanilla RAG 
& 0.058 & 0.339 & 0.068 & 58.968 
& 0.092 & 0.366 & 0.095 & 68.146 
& 0.027 \\

Mem0 
& 0.205 & 0.281 & 0.198 & 10.387
& 0.234 & 0.262 & 0.234 & 2.191 
& 0.036 \\

\midrule

\textbf{CraniMem (Ours)} 
& 0.315 & 0.317 & 0.312 & 112.440
& 0.329 & 0.325 & 0.323 & 53.977 
& 0.011 \\

\bottomrule
\end{tabular}
\caption{Comparison of different architectures on \textbf{Qwen2.5-7B-Instruct} under noisy and clean settings on multi-hop HotpotQA dataset.}
\label{tab:final-clean}
\end{table}

Table~\ref{tab:final-clean} shows that CraniMem performs overall better than Vanilla RAG and Mem0 under both clean and noisy settings, and exhibits the smallest noise drop. This indicates stronger robustness to distractors, with the main cost being higher latency due to CraniMem's more complex memory pipeline that includes the consolidation phase and storage through knowledge graph. 

\begin{table}[H]
\centering
\small
\renewcommand{\arraystretch}{1.08}
\setlength{\tabcolsep}{5pt}
\begin{tabular}{p{3.4cm}llccccc}
\hline
\textbf{Model} & \textbf{Architecture} & \textbf{Type} & \textbf{Precision} & \textbf{Recall} & \textbf{F1} &
\begin{tabular}[c]{@{}c@{}}
\textbf{Noise Drop}\\
\textbf{$\Delta_{\mathrm{noise}}$}
\end{tabular}
& \textbf{Latency (s)} \\
\hline
\multirow{4}{*}{\raggedright Qwen2.5-Coder-7B-Instruct}
 & \multirow{2}{*}{Vanilla RAG} & Noisy & 0.100 & 0.117 & 0.096 & \multirow{2}{*}{0.079} & 6.810 \\
 &  & Clean & 0.170 & 0.220 & 0.175 &  & 4.439 \\
 & \multirow{2}{*}{CraniMem} & Noisy & 0.289 & 0.279 & 0.280 &\multirow{2}{*}{0.004} & 252.915 \\
 &  & Clean & 0.294 & 0.282 & 0.284 &  & 9.029 \\
\hline
\multirow{4}{*}{\raggedright Gemma-2-9B-IT}
 & \multirow{2}{*}{Vanilla RAG} & Noisy & 0.123 & 0.116 & 0.116 & \multirow{2}{*}{0.160} & 61.023 \\
 &  & Clean & 0.303 & 0.268 & 0.276 &  & 44.448 \\
 & \multirow{2}{*}{CraniMem} & Noisy & \textbf{0.339} & \textbf{0.318} & \textbf{0.323} & \multirow{2}{*}{0.015} & 134.126 \\
 &  & Clean & 0.352 & 0.334 & 0.338 &  & 70.152 \\
\hline
\multirow{4}{*}{\raggedright Qwen2.5-7B-Instruct }
 & \multirow{2}{*}{Vanilla RAG} & Noisy & 0.058 & 0.339 & 0.068 & \multirow{2}{*}{0.027} & 58.968 \\
 &  & Clean & 0.092 & 0.366 & 0.095 &  & 68.146 \\
 & \multirow{2}{*}{CraniMem} & Noisy & 0.315 & 0.317 & 0.312 & \multirow{2}{*}{0.011} & \textbf{112.440} \\
 &  & Clean & 0.329 & 0.325 & 0.323 &  & 53.977 \\
\hline
\multirow{4}{*}{\raggedright Mistral-7B-Instruct-v0.3 }
 & \multirow{2}{*}{Vanilla RAG} & Noisy & 0.115 & 0.177 & 0.118 & \multirow{2}{*}{0.129} & 5.514 \\
 &  & Clean & 0.243 & 0.353 & 0.247 &  & 7.701 \\
 & \multirow{2}{*}{CraniMem} & Noisy & 0.300 & 0.300 & 0.294 & \multirow{2}{*}{0.022} & 381.230 \\
 &  & Clean & 0.320 & 0.325 & 0.316 &  & 13.943 \\
\hline
\end{tabular}
\caption{CraniMem benchmark performance on multi-hop HotpotQA dataset under injected noise, reported across LLM backbones.}
\label{tab:results-main}
\end{table}

Table~\ref{tab:results-main} shows that across multiple instruction-tuned models, CraniMem improves noisy condition performance and substantially reduces noise drop compared to Vanilla RAG. For example, with Gemma-2-9B-IT, CraniMem achieves the best noisy F1 of 0.323 and a noise drop of 0.015, while the Vanilla RAG baseline exhibits a larger degradation with noise drop 0.160. Similar patterns hold for Qwen2.5-Coder-7B-Instruct, where CraniMem has near zero noise drop 0.004 versus 0.079 for Vanilla RAG, indicating that goal directed gating prevents distractors from entering memory and protects retrieval quality. Qwen2.5-7B-Instruct yields the lowest CraniMem latency among the tested backbones at 112.44 seconds in the noisy setting, while Qwen2.5-Coder-7B-Instruct and Mistral-7B-Instruct-v0.3 incur higher overhead. These results suggest that CraniMem is most beneficial in settings where distraction resistance and information stability are prioritized over raw throughput.

\section{Limitations and Conclusion}

In this paper, we introduce Cranimem, a neurocognitively inspired agentic memory architecture that supports encoding, consolidation, and retrieval for long horizon agent behavior. We compare our architecture against baselines on a similar architecture of Mem0 and Vanilla RAG. CraniMem consistently outperforms both baselines, showing stronger robustness to distractors and more stable long horizon recall. This improved robustness comes with a deployment trade-off, CraniMem incurs higher latency. Due to computational constraints, our experiments were conducted on 100 sampled instances, which limits the statistical strength of the reported numbers. Future work will expand the empirical study to larger evaluation sets and perform systematic comparisons and ablations against other memory architectures, including HippoRAG and Mem0, under standardized engineering and evaluation settings. For reproducibility, we deployed our framework as a PyPI package and release the accompanying code and prompts, enabling reproduction of our results and comparable benchmarking against Vanilla RAG.





\bibliography{iclr2026_conference}
\bibliographystyle{iclr2026_conference}

\appendix

\section{Datasets} \label{datasets}
HotpotQA is a multi-hop question answering dataset that requires combining evidence across multiple supporting facts. We evaluate CraniMem on a sampled subset of HotpotQA dataset. 

We construct a sample of 100 instances from the HotpotQA validation split and use this subset to test the long horizon behavior of our memory pipeline, including selective write gating, bounded episodic storage, and consolidation into the semantic store.

\section{CraniMem Implementation Details}
\label{appendix:cranimem_details}

\subsection{Input gating and utility tagging}
Given the user message $u_t$ and the agent goal and state $g_t$, CraniMem computes semantic relevance using an embedding model $E(\cdot)$:
\begin{equation*}
\mathrm{Sim}(u_t,g_t) = \mathrm{CosineSimilarity}\big(E(u_t),E(g_t)\big).
\end{equation*}
If $\mathrm{Sim}(u_t,g_t) < \tau_{\mathrm{noise}}$, the input is discarded. Otherwise, an LLM-based tagger produces scalar signals for importance, surprise, and emotion, which are aggregated into a base utility score:
\begin{equation*}
\mathrm{BaseUtility}(u_t) = \tfrac{1}{3}\big(I(u_t) + S(u_t) + E(u_t)\big).
\end{equation*}

\subsection{Episodic buffer}
Accepted items are stored in a bounded FIFO episodic buffer over the most recent $N$ turns. Each entry contains the raw snippet, timestamp and turn id, extracted entities, and utility metadata. The buffer supports short-range continuity and provides the candidate set for consolidation.

\subsection{Replay selection and consolidation}
CraniMem triggers consolidation every $N$ turns or during idle states. For each buffer item $m$, a replay score combines intrinsic utility with a frequency bonus based on repeated entities and repeated access:
\begin{equation*}
\mathrm{ReplayScore}(m) = \mathrm{BaseUtility}(m)\,\Big(1 + \alpha\,\mathrm{FreqBonus}(m)\Big).
\end{equation*}
Only items with $\mathrm{ReplayScore}(m) > \tau_{\mathrm{consolidation}}$ are selected for transfer to long-term memory.

\subsection{Linkage engine and knowledge graph updates}
Selected memories are converted into a knowledge graph by extracting typed entities and typed relations, applying checks, and performing an upsert and merge operation that reinforces repeated facts. This representation supports multi-hop recall through graph traversal, which is difficult for flat vector stores.

\subsection{Dual-path retrieval}
At generation time, CraniMem retrieves evidence from the episodic buffer for short-range continuity and from the knowledge graph for long-range semantic recall. The two streams are merged into a compact context block that conditions the LLM.

\section{Evaluation Metrics} \label{evaluation_metrics}
We report both clean and noisy performance to quantify robustness of memory under context pollution.

\paragraph{Clean metrics:}
On a clean evaluation setting without injected distractors, we compute the averaged standard classification and retrieval quality measures of precision, recall, and F1 score.

\paragraph{Noisy metrics:}
To test distraction resistance, we evaluate under a noisy setting where irrelevant context is injected into the agent's input stream. We again report the averaged precision, recall, and F1 score.

\paragraph{Noise drop:}
We measure degradation due to noise using noise drop, defined as the gap between clean and noisy F1 score:
\begin{equation*}
\Delta_{\mathrm{noise}} = \mathrm{F1}_{\mathrm{clean}} - \mathrm{F1}_{\mathrm{noisy}}.
\end{equation*}
Lower noise drop indicates higher robustness and better memory gating and retrieval precision under distraction.

\paragraph{Latency:}
Finally, we report latency as the end-to-end per turn inference time, including memory read, memory write, and consolidation overhead when triggered. Latency captures the practical cost of memory operations and highlights the trade-off between robustness and runtime efficiency.

\section{Prompts}
\label{appendix:prompts}
This section reports the prompt templates used for goal directed gating, utility tagging, and knowledge graph extraction.

\begin{promptbox}{Gating System Prompt}
You are the goal directed Gating Mechanism.
Your job is to filter information based strictly on its RELEVANCE to the current Agent Goal.

### CURRENT AGENT GOAL:
"{goal}"

### INSTRUCTIONS:
Analyze the User Input. Does it contain information that helps achieve the Goal?

1. **HIGH PRIORITY (Score 8-10):**
   - Directly contributes to the goal (e.g., specific facts, constraints, deadlines).
   - "Project X is due Friday" (If goal is Project Management).
   - "I am allergic to peanuts" (If goal is Food Ordering).

2. **MEDIUM PRIORITY (Score 4-7):**
   - Context that *might* be useful later.
   - Clarifications or minor updates.

3. **NOISE (Score 0-3):**
   - Information irrelevant to the current goal.
   - Chitchat, greetings, or off-topic distractions.

...
IMPORTANT: Output ONLY raw JSON. Do not include any preamble, markdown, or explanations.
Output strictly in JSON format:
{{
  "is_noise": boolean, 
  "priority_score": integer, // SCALE: 1-10. 
                             // 10 = Critical technical facts, research data, or goal-aligned info.
                             // 7-9 = Relevant context or useful updates.
                             // 1-6 = Mundane chitchat, emotional noise, or irrelevant distractions.
  "entities": [list of strings],
  "reasoning": "string"
}}
\end{promptbox}

\begin{promptbox}{Utility Tagging System Prompt}
You are a Utility Tagger.
Analyze the input for three RAS factors and return JSON only.
IMPORTANT: Output ONLY raw JSON. Do not include any preamble, markdown, or explanations.
Output ONLY JSON. Do not write any conversational text, preambles, or explanations.

Factors:
1) Importance: Is it technical/goal-oriented?
2) Surprise: Is it new or unexpected?
3) Emotion: Is there strong sentiment?

Guidelines:
- Each factor is a float in [0, 1].
- If the input is clearly irrelevant or trivial, keep scores near 0.
- "entities" should be key proper nouns or concepts, if any.

Return JSON:
{{
  "importance": float,
  "surprise": float,
  "emotion": float,
  "entities": [list of strings]
}}
\end{promptbox}

\begin{promptbox}{Reflex Utility System Prompt}
"You are a Utility Tagger.
The vector match is HIGH, so the input is already considered relevant.
Return ONLY the utility scores.
IMPORTANT: Output ONLY raw JSON. Do not include any preamble, markdown, or explanations.

Return JSON:
{{
  "importance": float,
  "surprise": float,
  "emotion": float,
  "entities": [list of strings]
}}
\end{promptbox}

\begin{promptbox}{Cortex Gating System Prompt}
You are the Cortex Gate for low vector similarity inputs.
The vector match is LOW, so you must decide whether this input is:
1) a command, 
2) a goal change / context shift, 
3) relevant context, or 
4) noise.

IMPORTANT: Output ONLY raw JSON. Do not include any preamble, markdown, or explanations.

Return JSON:
{{
  "is_noise": boolean,
  "category": "command|goal_change|relevant_context|noise",
  "importance": float,
  "surprise": float,
  "emotion": float,
  "entities": [list of strings],
  "reason": "short string"
}}
\end{promptbox}

\begin{promptbox}{Reasoning Prompt}
You are a precise answering machine.
GOAL: "{current_goal}"

CONTEXT:
{context}

USER INPUT:
{current_input}

INSTRUCTIONS:
1. Answer the question accurately using the context.
2. Output ONLY the answer entity (Name, Place, Date, etc.).
3. NO full sentences. NO "The answer is...".
4. If the answer is a long name (e.g., "The United States of America"), write the FULL name.
5. Wrap your final answer inside <RESPONSE> tags.

Example:
User: Who directed the movie?
<RESPONSE>Scott Derrickson</RESPONSE>

User: What organization?
<RESPONSE>International Boxing Hall of Fame</RESPONSE>

YOUR TURN:
\end{promptbox}

\begin{promptbox}{Entity Relation Extraction System Prompt }
You are a Knowledge Graph extraction system.
Extract only the entities and relationships that are explicitly supported by the text.

STRICT FILTERING:
1) Ignore general knowledge or common sense statements (e.g., "The sky is blue", "Pizza is tasty", "Python is a language").
2) Focus ONLY on specific, unique facts about people, places, organizations, events, and dates.
3) Prefer dense/complex information: multi-entity, multi-relation, or highly specific facts.
4) If a sentence contains no specific named entities, SKIP IT.

IMPORTANT: Output ONLY raw JSON. Do not include any preamble, markdown, or explanations.

Return JSON only, with this schema:
{{
  "entities": [
    {{"type": "Project|Issue|Task|Person|Tool|Feature|Location|Date|Other", "name": "string"}}
  ],
  "relations": [
    {{"source": "entity name", "relation": "string", "target": "entity name"}}
  ]
}}

Rules:
1) If no entities exist, return empty lists.
2) Do not invent entities or relations.
3) Use concise names, no labels in the name field.
4) relation should be a short verb phrase (e.g., "has_issue", "blocked_by", "uses").
\end{promptbox}

\begin{promptbox}{Entity Extraction Prompt}
You are a Knowledge Graph Engineer.
    Extract the key "Entities" (Proper Nouns, Project Names, Locations, Dates) from the text below.
    
    Rules:
    1. Return ONLY a comma-separated list of entities.
    2. Do not include labels (e.g., don't say "Person: Sarah", just say "Sarah").
    3. If no entities are found, return "None".
    
    TEXT: {input}
    
    ENTITIES:
\end{promptbox}

\end{document}